# Sparse Nested Markov Models with Log-linear Parameters

**Ilya Shpitser**
Mathematical Sciences
University of Southampton
i.shpitser@soton.ac.uk

**Robin J. Evans**
Statistical Laboratory
Cambridge University
rje42@cam.ac.uk

**Thomas S. Richardson**
Department of Statistics
University of Washington
thomasr@u.washington.edu

**James M. Robins**
Department of Epidemiology
Harvard University
robins@hsph.harvard.edu

## Abstract

Hidden variables are ubiquitous in practical data analysis, and therefore modeling marginal densities and doing inference with the resulting models is an important problem in statistics, machine learning, and causal inference. Recently, a new type of graphical model, called the nested Markov model, was developed which captures equality constraints found in marginals of directed acyclic graph (DAG) models. Some of these constraints, such as the so called 'Verma constraint', strictly generalize conditional independence. To make modeling and inference with nested Markov models practical, it is necessary to limit the number of parameters in the model, while still correctly capturing the constraints in the marginal of a DAG model. Placing such limits is similar in spirit to sparsity methods for undirected graphical models, and regression models. In this paper, we give a log-linear parameterization which allows sparse modeling with nested Markov models. We illustrate the advantages of this parameterization with a simulation study.

## 1 Introduction

Analysis of complex multidimensional data is often made difficult by the twin problems of hidden variables, and a dearth of data relative to the dimension of the model. The former problem motivates the study of marginal and/or latent models, while the latter has resulted in the development of sparsity methods.

A particularly appealing model for multidimensional data analysis is the Bayesian network or directed acyclic graph (DAG) model [10], where random variables are represented as vertices in the graph, with directed edges (arrows) between them. The popularity of DAG models stems from their well understood theory, and from the fact that they elicit an intuitive causal interpretation: an arrow from a variable $A$ to a variable $B$ in a DAG model can be interpreted, in a way which can be made precise, to mean that $A$ is a 'direct cause' of $B$.

DAG models assume all variables are observed, and a latent variable model based on DAGs simply relaxes this assumption. However, latent variables introduce a number of problems: it is difficult to correctly model the latent state, and the resulting marginal densities are quite challenging to work with. An alternative is to encode constraints found in marginals of DAG models directly; a recent approach in this spirit is the nested Markov model [15]. The advantage of the nested Markov model is that it correctly captures the conditional independences and other equality constraints found in marginals of DAG models. However, the discrete parameterization of nested Markov models has the disadvantage of being unable to represent constraints in various marginals of DAGs *concisely*, that is with few non-zero parameters. This implies that model selection methods based on scoring (via the BIC score [13] for instance) often prefer simpler models which fail to capture independences correctly, but which contain many fewer parameters [15].

More generally, in high dimensional data analyses there is often such a shortage of samples that classical statistical inference techniques do not work. To address these issues, sparsity methods have been developed, which drive as many parameters in the statistical model to zero as possible, while still providing a reasonable fit to the data. Sparsity methods have been developed for regression models [16], undirected graphical models [8, 9], and even some marginal models [4].

It is not natural to apply sparsity techniques to existing parameterizations of nested Markov models, because the parameters are context (or strata) specific.

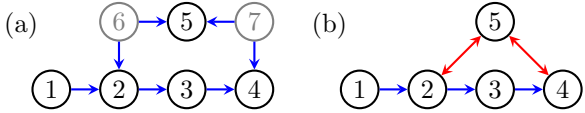

Figure 1: (a) A DAG with nodes 6 and 7 representing hidden variables. (b) An ADMG representing the same conditional independences as (a) among the variables corresponding to 1, 2, 3, 4, 5.

In this paper, we develop a log-linear parameterization for discrete nested Markov models, where the parameters represent (generalizations of) log odds-ratios within 'kernels' (informally 'interventional' densities). These can be viewed as interaction parameters, of the kind commonly set to zero by sparsity methods. Our parameterization allows us to represent distributions containing 'Verma constraints' in a sparse way, while maintaining advantages of nested Markov models, and avoiding the disadvantages of using marginals of DAG models directly.

## 2 Disadvantages of the Möbius Parameterization of Nested Markov Models

One drawback of the standard parameterization of nested Markov models is that parameters are variation dependent; that is, fixing the value of one parameter constrains the 'legal' values of other parameters. This is in direct contrast with parameterizations of DAG models where parameters associated with a particular Markov factor (a conditional density for a variable given all its parents in the DAG) do not depend on parameters associated with other Markov factors.

We illustrate another difficulty with an example. Here, and in subsequent discussions, we will need to draw distinctions between vertices in graphs, and corresponding random variables in distributions or 'kernels.' We use the following notation: $v$ (lowercase) denotes a vertex, $X_v$ the corresponding random variable, and $x_v$ a value assignment to this variable. Likewise $A$ (uppercase) denotes a vertex set, $X_A$ the corresponding random variable set, and $x_A$ an assignment to this set.

Consider the marginal DAG shown in Fig. 1 (a). We wish to avoid representing this domain with a DAG directly, in order not to commit to a particular state space of the unobserved variables $X_6$ and $X_7$, and because, even if we were willing to make such an assumption, the margin over $(X_1, X_2, X_3, X_4, X_5)$ obtained from a density that factorizes according to this DAG can be complicated to work with [7].

To use nested Markov models for this domain, we first construct an acyclic directed mixed graph (ADMG) that represents this DAG marginal, using the latent projection algorithm [17]. This graph is shown in Fig. 1 (b); directed arrows in the resulting ADMG represent directed paths in the DAG where any intermediate nodes are unobserved (in this case there are no such paths, and all directed edges in the ADMG are directly inherited from the DAG). Similarly, bidirected arrows in the ADMG, such as $2 \leftrightarrow 5$, represent marginally d-connected paths in the DAG which start and end with arrowheads pointing away from the path, in this case $2 \leftarrow 6 \rightarrow 5$.

If we now use the nested Möbius parameters, described in more detail in subsequent sections, to parameterize the resulting ADMG, we will quickly discover that this results in a model of higher dimension relative to the dimension of DAG models which share their skeleton with this ADMG. For example, the binary nested Markov model of the graph in Fig. 1 (b) has 16 parameters, while both binary DAG models corresponding to graphs in Fig. 7 (a) and (b) have 11 parameters each.

This leads to a worry that a structure learning algorithm that tries to use nested Möbius parameters to recover an ADMG from data by means of a score method, such as BIC [13], which rewards fit and parameter parsimony, may prefer at low sample sizes incorrect independence models given by DAGs in preference to correct models given by ADMGs, simply because the DAG models compensate for their poor fit of the data with a much smaller parameter count. In fact, this precise issue has been observed in simulation studies reported in [15].

Addressing this problem with a Möbius parameterization is not easy, because Möbius parameters are strata or context-specific; in other words, the parameterization is not independent of how the states are labeled. For instance, some of the Möbius parameters representing confounding between $X_2, X_4$ and $X_5$ are: [1]

$$\theta_{\{2,4,5\}}(x_1, x_3) = p(0_4, 0_5 | x_3, 0_2, x_1)\, p(0_2 | x_1)$$

for all values of $x_1, x_3$. In a binary model, this gives 4 parameters. The kinds of regularities in the true generative process, which we may want to exploit to create a dimension reduction in our model, typically involve a lack of interactions among variables, or a latent confounder with a low dimensional state space. Such regularities may often not translate into constraints naturally expressible in terms of Möbius parameters.

To avoid this difficulty, we need to construct parameters for nested Markov models which represent various types of interactions among variables directly. In fact, parameters representing interactions are well known in log-linear models, of which undirected graphical models and certain regression models form a special case.

---
[1]To save space, here and elsewhere we will write $1_i$ for an assignment of $X_i$ to 1, and $0_i$ for an assignment to 0.

## 3 Log-linear Parameters for Undirected Models

We will use undirected graphical models, also known as Markov random fields, to illustrate log-linear models. A Markov random field over a multivariate binary state space $\mathfrak{X}_V$, is a set of densities $p(x_V)$ represented by an undirected graph $\mathcal{G}$ with vertices $V$, where

$$p(x_V) = \exp\left(\sum_{C \in \mathrm{cl}(\mathcal{G})} (-1)^{\|x_C\|_1} \lambda_C\right);$$

here $\mathrm{cl}(\mathcal{G})$ is the collection of (not necessarily maximal) cliques in the undirected graph, $\|\cdot\|_1$ is the $L_1$-norm, and $\lambda_C$ is a *log-linear parameter*. Note that the parameter $\lambda_\emptyset$ ensures the expression is normalized.

Consider the undirected graph shown in Fig. 2. In this graph, all subsets of $\{1,2,3\}$, $\{2,4\}$, and $\{4,5,6\}$ are cliques. The model represents densities where, conditional upon its adjacent nodes, each node is independent of all others. The log-linear parameter(s) corresponding to each such subset of size $k$ can be viewed as representing $k$-way interactions among appropriate variables in the model. Setting some such interaction parameters to zero in a consistent way results in a model which still asserts the same conditional independences, but has a smaller parameter count, and with all strata in each clique treated symmetrically. For instance, if we were to set all parameters for cliques of size $k > 2$ to zero, so that there remained only parameters corresponding to vertices and individual edges, we would obtain a model known as a Boltzmann machine [1]. A similar idea had been used to give a sparse parameterization for discrete DAG models [12].

In the remainder of the paper, we describe nested Markov models, and give a log-linear parameterization for these models which contains similar parameters that may be set to zero. While in Markov random field models the parameters are associated with sets of nodes which form cliques in the corresponding undirected graph, in nested Markov models parameters will be associated with special sets of nodes in the corresponding ADMG called *intrinsic sets*. Further, log-linear parameterizations of this type can often incorporate individual-level continuous baseline covariates [5].

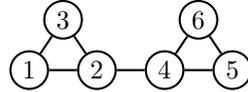

Figure 2: An undirected graph representing a log-linear model.

## 4 Graphs, Kernels, and Nested Markov Models

We now introduce the relevant background needed to define the nested Markov model.

A *directed mixed graph* $\mathcal{G}(V,E)$ is a graph with a set of vertices $V$ and a set of edges $E$, where the edges may be directed ($\to$) or bidirected ($\leftrightarrow$). A directed cycle is a path of the form $x \to \cdots \to y$ along with an edge $y \to x$. An *acyclic* directed mixed graph (ADMG) is a mixed graph containing no directed cycles. An example is given in Fig. 1 (b).

Let $a$, $b$ and $d$ be vertices in a mixed graph $\mathcal{G}$. If $b \to a$ then we say that $b$ is a *parent* of $a$, and $a$ is a *child* of $b$. If $a \leftrightarrow b$ then $a$ is said to be a *spouse* of $b$. A vertex $a$ is said to be an *ancestor* of a vertex $d$ if *either* there is a directed path $a \to \cdots \to d$ from $a$ to $d$, *or* $a = d$; similarly $d$ is said to be a *descendant* of $a$. The sets of parents, children, spouses, ancestors and descendants of $a$ in $\mathcal{G}$ are written $\mathrm{pa}_\mathcal{G}(a)$, $\mathrm{ch}_\mathcal{G}(a)$, $\mathrm{sp}_\mathcal{G}(a)$, $\mathrm{an}_\mathcal{G}(a)$, $\mathrm{de}_\mathcal{G}(a)$ respectively. We apply these definitions disjunctively to sets, e.g. $\mathrm{an}_\mathcal{G}(A) = \bigcup_{a \in A} \mathrm{an}_\mathcal{G}(a)$.

### 4.1 Conditional ADMGs

A *conditional* acyclic directed mixed graph (CADMG) $\mathcal{G}(V,W,E)$ is an ADMG with a vertex set $V \cup W$, where $V \cap W = \emptyset$, subject to the restriction that for all $w \in W$, $\mathrm{pa}_\mathcal{G}(w) = \emptyset = \mathrm{sp}_\mathcal{G}(w)$. The vertices in $V$ are the *random* vertices, and those in $W$ are called *fixed*.

Whereas an ADMG with vertex set $V$ represents a joint density $p(x_V)$, a conditional ADMG represents the Markov structure of a conditional density, or kernel $q_V(x_V|x_W)$. Following [8, p.46], we define a *kernel* to be a non-negative function $q_V(x_V|x_W)$ satisfying:

$$\sum_{x_V} q_V(x_V \mid x_W) = 1 \qquad \text{for all } x_W. \qquad (1)$$

We use the term 'kernel' and write $q_V(\cdot|\cdot)$ (rather than $p(\cdot|\cdot)$) to emphasize that these functions, though they satisfy (1) and thus most properties of conditional densities, are not in general formed via the usual operation of conditioning on the event $X_W = x_W$. To conform with standard notation for densities, for every $A \subseteq V$

let

$$q_V(x_A|x_W) \equiv \sum_{V \setminus A} q_V(x_V|x_W),$$

$$q_V(x_{V \setminus A}|x_{W \cup A}) \equiv \frac{q_V(x_V|x_W)}{q_V(x_A|x_W)}.$$

For a CADMG $\mathcal{G}(V, W, E)$ we consider collections of random variables $(X_v)_{v \in V}$ indexed by variables $(X_w)_{w \in W}$; throughout this paper the random variables take values in finite discrete sets $(\mathfrak{X}_v)_{v \in V}$ and $(\mathfrak{X}_w)_{w \in W}$. For $A \subseteq V \cup W$ we let $\mathfrak{X}_A \equiv \times_{u \in A}(\mathfrak{X}_u)$, and $X_A \equiv (X_v)_{v \in A}$. That we will always hold the variables in $W$ fixed is precisely why we do not permit edges between vertices in $W$.

An ADMG $\mathcal{G}(V, E)$ may be seen as a CADMG in which $W = \emptyset$. In this manner, though we will state subsequent definitions for CADMGs, they also apply to ADMGs.

The *induced subgraph* of a CADMG $\mathcal{G}(V, W, E)$ given by a set $A$, denoted $\mathcal{G}_A$, consists of $\mathcal{G}(V \cap A, W \cap A, E_A)$ where $E_A$ is the set of edges in $\mathcal{G}$ with both endpoints in $A$. In forming $\mathcal{G}_A$, the status of the vertices in $A$ with regard to whether they are in $V$ or $W$ is preserved.

### 4.2 Districts and Markov Blankets

A set $C$ is *connected* in $\mathcal{G}$ if every pair of vertices in $C$ is joined by a path such that every vertex on the path is in $C$. For a given CADMG $\mathcal{G}(V, W, E)$, denote by $(\mathcal{G})_\leftrightarrow$ the CADMG formed by removing all directed edges from $\mathcal{G}$. A set connected in $(\mathcal{G})_\leftrightarrow$ is called *bidirected connected*.

For a vertex $x \in V$, the *district* (or c-component) of $x$, denoted by $\text{dis}_\mathcal{G}(x)$, is the maximal bidirected connected set containing $x$. For instance in the ADMG shown in Fig. 1 (b), the district of node 2 is $\{2, 4, 5\}$. Districts in a CADMG form a partition of $V$; vertices in $W$ are excluded by definition. In a DAG $\mathcal{G}(V, E)$ the set of districts is the set of all single element sets $\{v\} \subseteq V$.

A set of vertices $A$ in $\mathcal{G}$ is called *ancestral* if $a \in A \Rightarrow \text{an}_\mathcal{G}(a) \subseteq A$. In a CADMG $\mathcal{G}(V, W, E)$, if $A$ is an ancestral subset of $V \cup W$ in $\mathcal{G}$, $t \in A \cap V$, and $\text{ch}_\mathcal{G}(t) \cap A = \emptyset$, then the *Markov blanket of $t$ in $A$* is defined as:

$$\text{mb}_\mathcal{G}(t, A) \equiv \text{pa}_\mathcal{G}\Big(\text{dis}_{\mathcal{G}_A}(t)\Big) \cup \Big(\text{dis}_{\mathcal{G}_A}(t) \setminus \{t\}\Big).$$

### 4.3 The fixing operation and fixable vertices

We now introduce a 'fixing' operation on a CADMG which has the effect of transforming a random vertex into a fixed vertex, thereby changing the graph. However, this operation is only applicable to a subset of the vertices, which we term the (potentially) fixable vertices.

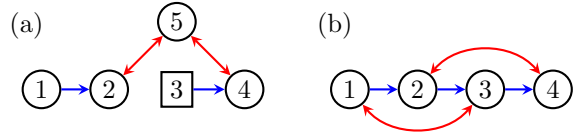

Figure 3: (a) The graph from Fig. 1 (b) after fixing 3. (b) An ADMG inducing a non-trivial nested Markov model.

**Definition 1** *Given a CADMG $\mathcal{G}(V, W, E)$ the set of fixable vertices is*

$$\mathbb{F}(\mathcal{G}) \equiv \{v \mid v \in V, \text{dis}_\mathcal{G}(v) \cap \text{de}_\mathcal{G}(v) = \{v\}\}.$$

In words, $v$ is fixable in $\mathcal{G}$ if there is no vertex $v^*$ that is both a descendant of $v$ and in the same district as $v$. For the graph in Fig. 1 (b), the vertex 2 is not fixable, because 4 is both its descendant and in the same district; all the other vertices are fixable.

**Definition 2** *Given a CADMG $\mathcal{G}(V, W, E)$, and a kernel $q_V(X_V \mid X_W)$, with every $r \in \mathbb{F}(\mathcal{G})$ we associate a fixing transformation $\phi_r$ on the pair $(\mathcal{G}, q_V(X_V \mid X_W))$ defined as follows:*

$$\phi_r(\mathcal{G}) \equiv \mathcal{G}^*(V \setminus \{r\}, W \cup \{r\}, E_r),$$

*where $E_r$ is the subset of edges in $E$ that do not have arrowheads into $r$, and*

$$\phi_r(q_V(x_V \mid x_W); \mathcal{G}) \equiv \frac{q_V(x_V \mid x_W)}{q_V(x_r \mid x_{\text{mb}_\mathcal{G}(r, \text{an}_\mathcal{G}(\text{dis}_\mathcal{G}(r)))})}$$

Returning to the ADMG in Fig. 1 (b), fixing 3 in the graph means removing the edge $2 \rightarrow 3$, while fixing $x_3$ in $p(x_1, x_2, x_3, x_4, x_5)$ means dividing this marginal density by $q_{1,2,3,4,5}(x_3 \mid x_2) = p(x_3 \mid x_2)$. The resulting CADMG, shown in Fig. 3 (a), represents the resulting kernel $q_{1,2,4,5}(x_1, x_2, x_4, x_5 \mid x_3)$.

We use $\circ$ to indicate composition of operations in the natural way, so that: $\phi_r \circ \phi_s(\mathcal{G}) \equiv \phi_r(\phi_s(\mathcal{G}))$ and

$$\phi_r \circ \phi_s(q_V(X_V|X_W); \mathcal{G})$$
$$\equiv \phi_r(\phi_s(q_V(X_V|X_W); \mathcal{G}); \phi_s(\mathcal{G})).$$

### 4.4 Reachable and Intrinsic Sets

In order to define the nested Markov model, we will need to define special classes of vertex sets in ADMGs.

**Definition 3** *A CADMG $\mathcal{G}(V,W)$ is* reachable *from an ADMG $\mathcal{G}^*(V \cup W)$ if there is an ordering of the vertices in $W = \langle w_1, \ldots, w_k \rangle$, such that for $j = 1, \ldots, k$,*

$$w_1 \in \mathbb{F}(\mathcal{G}^*) \text{ and for } j = 2, \ldots, k,$$
$$w_j \in \mathbb{F}(\phi_{w_{j-1}} \circ \cdots \circ \phi_{w_1}(\mathcal{G}^*)).$$

A subgraph is reachable if, under some ordering, each vertex $w_i$ is fixable in $\phi_{w_{i-1}}(\cdots \phi_{w_2}(\phi_{w_1}(\mathcal{G}^*)) \cdots)$.

Fixing operations do not in general commute, and thus only some orderings are valid for fixing a particular set. For example, in the ADMG shown in Fig. 1 (b), the set $\{2,3\}$ may be fixed, but only under the ordering where 3 is fixed first, to yield the CADMG shown in Fig. 3 (a), and then 2 is fixed in this CADMG. Fixing 2 first in Fig. 1 (b) is not valid, because 4 is both a descendant of 2 and in the same district as 2 in that graph, and thus 2 is not fixable.

If a CADMG $\mathcal{G}(V,W)$ is reachable from $\mathcal{G}^*(V \cup W)$, we say that the set $V$ is reachable in $\mathcal{G}^*$. A reachable set may be obtained by fixing vertices using more than one valid sequence. We will denote any valid composition of fixing operations that fixes a set $A$ by $\phi_A$ if applied to the graph, and by $\phi_{X_A}$ if applied to a kernel. With a slight abuse of notation (though justified as we will later see) we suppress the precise fixing sequence chosen.

**Definition 4** *A set of vertices $S$ is* intrinsic *in $\mathcal{G}$ if it is a district in a reachable subgraph of $\mathcal{G}$. The set of intrinsic sets in an ADMG $\mathcal{G}$ is denoted by $\mathcal{I}(\mathcal{G})$.*

For example, in the graph in Fig. 1 (b), the set $\{2,4,5\}$ is intrinsic (and reachable), while the set $\{1,2,4,5\}$ is reachable but not intrinsic.

In any DAG $\mathcal{G}(V,E)$, $\mathcal{I}(\mathcal{G}) = \{\{x\} | x \in V\}$, while in any bidirected graph $\mathcal{G}$, $\mathcal{I}(\mathcal{G})$ is equal to the set of all connected sets in $\mathcal{G}$.

### 4.5 Nested Markov Models

Just as for DAG models, nested Markov models may be defined via one of several equivalent Markov properties. These properties are all *nested* in the sense that they apply recursively to either reachable or intrinsic sets derived from an ADMG. In particular, there is a nested analogue of the global Markov property for DAGs (d-separation), the local Markov property for DAGs (which asserts that variables are independent of non-descendants given parents), and the moralization-based property for DAGs. These definitions appear and are proven equivalent in [11]. It is possible to associate a unique ADMG with a particular marginal DAG model, and a nested Markov model associated with this ADMG will recover all independences which hold in the marginal DAG [11].

We now define a nested factorization on probability distributions represented by ADMGs using special sets of nodes called 'recursive heads' and 'tails.'

**Definition 5** *For an intrinsic set $S \in \mathcal{I}(\mathcal{G})$ of a CADMG $\mathcal{G}$, define the recursive head (rh) as:* $\operatorname{rh}(S) \equiv \{x \in S \mid \operatorname{ch}_\mathcal{G}(x) \cap S = \emptyset\}$.

**Definition 6** *The* tail *associated with a recursive head $H$ of an intrinsic set $S$ in a CADMG $\mathcal{G}$ is given by:* $\operatorname{tail}(H) \equiv (S \setminus H) \cup \operatorname{pa}_\mathcal{G}(S)$.

In the graph in Fig. 1 (b), the recursive head of the intrinsic set $\{2,4,5\}$ is equal to the set itself, while the tail is $\{1,3\}$.

A kernel $q_V(X_V | X_W)$ satisfies the *head factorization property* for a CADMG $\mathcal{G}(V,W,E)$ if there exist kernels $\{f_S(X_H | X_{\operatorname{tail}(H)}) \mid S \in \mathcal{I}(\mathcal{G}), H = \operatorname{rh}_\mathcal{G}(S)\}$ such that

$$q_V(X_V | X_W) = \prod_{\substack{H \in [\![V]\!]_\mathcal{G} \\ S : \operatorname{rh}_\mathcal{G}(S) = H}} f_S(X_H | X_{\operatorname{tail}(H)}) \qquad (2)$$

where $[\![V]\!]_\mathcal{G}$ is a partition of $V$ into heads given in [14].

Let $\mathbb{G}(\mathcal{G}) \equiv \{(\mathcal{G}^*, \mathbf{w}^*) \mid \mathcal{G}^* = \phi_{\mathbf{w}^*}(\mathcal{G})\}$ for an ADMG $\mathcal{G}$. That is, $\mathbb{G}(\mathcal{G})$ is the set of valid fixing sequences and the CADMGs that they reach. The same graph may be reached by more than one sequence $\mathbf{w}^*$. We say that a distribution $p(x_V)$ obeys the *nested head factorization property* for $\mathcal{G}$ if for all $(\mathcal{G}^*, \mathbf{w}^*) \in \mathbb{G}(\mathcal{G})$, the kernel $\phi_{\mathbf{w}^*}(p(X_V); \mathcal{G})$ obeys the head factorization for $\phi_{\mathbf{w}^*}(\mathcal{G}) \equiv \mathcal{G}^*$. We denote the set of such distributions by $\mathcal{P}_h^n(\mathcal{G})$. Nested Markov models have been defined via a nested district factorization criterion [15], and a number of Markov properties [11]. The head factorization is another way of defining the nested Markov model due to the following result.

**Theorem 7** *The set $\mathcal{P}_h^n(\mathcal{G})$ is the nested Markov model of $\mathcal{G}$.*

Our decision to suppress the precise fixing sequence from the fixing operation applied to sets is justified, due to the following result.

**Theorem 8** *If $p(x_V)$ is in the nested Markov model of $\mathcal{G}$, then for any reachable set $A$ in $\mathcal{G}$, any valid fixing sequence on $V \setminus A$ gives the same CADMG over $A$, and the same kernel $q_A(x_A | x_{V \setminus A})$ obtained from $p(x_V)$.*

### 4.6 A Möbius Parameterization of Binary Nested Markov Models

We now give the original parameterization for binary nested Markov models. The approach generalizes in a straightforward way to finite discrete state spaces. Multivariate binary distributions in the nested Markov model for an ADMG $\mathcal{G}$ may be parameterized by the following:

**Definition 9** *The* nested Möbius parameters *associated with a CADMG $\mathcal{G}$ are a set of functions:* $\mathfrak{Q}_\mathcal{G} \equiv \{q_S(X_H = \mathbf{0} \,|\, x_{\text{tail}(H)})$ *for* $H = \text{rh}(S), S \in \mathcal{I}(\mathcal{G})\}$.

Intuitively, a parameter $q_S(X_H = \mathbf{0}|x_{\text{tail}(H)})$ is the probability that the variable set $X_H$ assumes values $\mathbf{0}$ in a kernel obtained from $p(x_V)$ by fixing $X_{V\setminus S}$, and conditioning on $X_{\text{tail}(H)}$. As a shorthand, we will denote the parameter $q_S(X_H = \mathbf{0}|x_{\text{tail}(H)})$ by $\theta_H(x_{\text{tail}(H)})$.

**Definition 10** *Let* $\nu : V \cup W \mapsto \{0,1\}$ *be an assignment of values to the variables indexed by* $V \cup W$. *Define* $\nu(T)$ *to be the values assigned to variables indexed by a subset* $T \subseteq V \cup W$. *Let* $\nu^{-1}(0) = \{v \mid v \in V, \nu(v) = 0\}$.

*A distribution* $P(X_V \mid X_W)$ *is said to be* parameterized *by the set* $\mathfrak{Q}_\mathcal{G}$, *for CADMG $\mathcal{G}$ if:*

$$p(X_V = \nu(V) \,|\, X_W = \nu(W)) = \sum_{B \,:\, \nu^{-1}(0) \cap V \subseteq B \subseteq V} (-1)^{|B \setminus \nu^{-1}(0)|} \times \prod_{H \in [\![B]\!]_\mathcal{G}} \theta_H(X_{\text{tail}(H)} = \nu(\text{tail}(H))) \quad (3)$$

*where the empty product is defined to be 1.*

For example, the graph shown in Fig. 3 (b) representing a model over binary random variables $X_1, X_2, X_3, X_4$ is parameterized by the following sets of parameters:

$$\theta_1 = p(0_1)$$
$$\theta_2(x_1) = p(0_2|x_1)$$
$$\theta_{1,3}(x_2) = p(0_3|x_2, 0_1)p(0_1)$$
$$\theta_3(x_2) = \sum_{x_1} p(0_3|x_2, x_1)p(x_1)$$
$$\theta_{2,4}(x_1, x_3) = p(0_4|x_3, 0_2, x_1)p(0_2|x_1)$$
$$\theta_4(x_3) = \sum_{x_2} p(0_4|x_3, x_2, x_1)p(x_2|x_1).$$

The total number of parameters is $1+2+2+2+4+2 = 13$, which is 2 fewer than a saturated parameterization of a 4 node binary model, which contains $2^4 - 1 = 15$ parameters. The two missing parameters reflect the fact that $\theta_4(x_3)$ does not depend on $x_1$, which is a constraint induced by the absence of the edge from 1 to 4 in Fig. 3 (b). Note that this constraint is not a conditional independence. In fact, no conditional independences over variables corresponding to vertices $1, 2, 3, 4$ are advertised in Fig. 3 (b).

This parameterization maps $\theta_H$ parameters to probabilities in a CADMG via the inverse Möbius transform given by (3), and generalizes both the standard Markov parameterization of DAGs in terms of parameters of the form $p(x_i = 0 \,|\, \text{pa}(x_i))$, and the parameterization of bidirected graph models given in [3].

## 5 A Log-linear Parameterization of Nested Markov Models

We begin by defining a set of objects which are functions of the observed density, and which will serve as our parameters.

**Definition 11** *Let $\mathcal{G}(V,E)$ be an ADMG and $p(x_V)$ a density over a set of binary random variables $X_V$ in the nested Markov model of $\mathcal{G}$. For any $S \in \mathcal{I}(\mathcal{G})$, let $M = S \cup \text{pa}_\mathcal{G}(S)$, $A \subseteq M$ (with $A \cap S \neq \emptyset$), and let $q_S(x_S|x_{M\setminus S}) = \phi_{V\setminus S}(p(x_V); \mathcal{G})$ be the associated kernel. Then define*

$$\lambda_A^M = \frac{1}{2^{|M|}} \sum_{x_M} (-1)^{\|x_A\|_1} \log q_S(x_S|x_{M\setminus S}),$$

*to be the* nested log-linear parameter *associated with $A$ in $S$. Further let $\Lambda(\mathcal{G})$ be the collection*

$$\{\lambda_A^M \mid S \in \mathcal{I}(\mathcal{G}), M = S \cup \text{pa}_\mathcal{G}(S), \text{rh}_\mathcal{G}(S) \subseteq A \subseteq M\}$$

*of these log-linear parameters. We call $\Lambda(\mathcal{G})$ the* nested ingenuous parameterization *of $\mathcal{G}$.*

This parameterization is based on the graphical concepts of recursive heads and corresponding tails. We call the parameterization 'nested ingenuous' due to its similarity to a marginal log-linear parameterization called ingenuous in [6], and in contrast to other log-linear parameterizations which may exist for nested Markov models. Marginal model parameterizations of this type were first introduced in [2]. This definition extends easily to non-binary discrete data, in which case some parameters $\lambda_A^M$ become collections of parameters.

As an example, consider the graph shown in Fig. 3 (b) which represents a binary nested Markov model. The nested ingenuous parameters associated with the marginal $p(x_1)$ and conditional $p(x_2|x_1)$ are

$$\lambda_1^1 = \frac{1}{2} \log \frac{p(0_1)}{p(1_1)}$$
$$\lambda_2^{21} = \frac{1}{4} \log \frac{p(0_2|0_1) \cdot p(0_2|1_1)}{p(1_2|0_1) \cdot p(1_2|1_1)}$$
$$\lambda_{21}^{21} = \frac{1}{4} \log \frac{p(0_2|0_1) \cdot p(1_2|1_1)}{p(1_2|0_1) \cdot p(0_2|1_1)}$$

whereas parameters associated with the kernel
$q_4(x_4|x_3) = \sum_{x_2} p(x_4|x_3, x_2, x_1) p(x_2|x_1)$ are

$$\lambda_4^{43} = \frac{1}{4} \log \frac{q_4(0_4|0_3) \cdot q_4(0_4|1_3)}{q_4(1_4|0_3) \cdot q_4(1_4|1_3)}$$

$$\lambda_{43}^{43} = \frac{1}{4} \log \frac{q_4(0_4|0_3) \cdot q_4(1_4|1_3)}{q_4(1_4|0_3) \cdot q_4(0_4|1_3)}$$

A parameter $\lambda_A^M$, where $M$ is the union of a head $H$ and its tail $T$, can be viewed, by analogy with similar clique parameters in undirected log-linear models, as a $|A|$-way interaction between the vertices in $A$, within the kernel corresponding to $M$. For instance the kernel $q_{2,4}(x_2, x_4|x_1, x_3) = p(x_4|x_3, x_2, x_1) p(x_2|x_1)$,[2] makes an appearance in 4 parameters in a binary model: $\lambda_{24}^{1234}$, $\lambda_{124}^{1234}$, $\lambda_{234}^{1234}$, and $\lambda_{1234}^{1234}$. If we set $\lambda_{1234}^{1234}$ to $0$, we claim there is no 4-way interaction between $X_1, X_2, X_3, X_4$ in the kernel.

It can be shown that while the Möbius parameterization of the graph in Fig. 3 (b) is variation dependent, the nested ingenuous parameterization of the same graph is variation independent. This is not true in general. In particular both parameterizations for the graph in Fig. 1 (b) are variation dependent.

## 6 Main Results

In this section we prove that the nested ingenuous parameters indeed parameterize discrete nested Markov models. We start with an intermediate result.

**Lemma 12** *Let $H \subseteq M$ and $q(x_H | x_{M \setminus H})$ be a kernel. Then $q$ is smoothly parameterized by the collection of NLL parameters $\{\lambda_A^M | H \subseteq A \subseteq M\}$ together with the $(|H|-1)$-dimensional margins of $q$, $q(x_{H \setminus \{v\}} | x_{M \setminus H}), v \in H$.*

*Proof:* The proof is essentially identical to the proof of Lemma 4.4 in [6]. □

### 6.1 The Main Result

We now define a partial order on heads and use this order to inductively establish the main result.

**Definition 13** *Let $\prec_{\mathcal{I}(\mathcal{G})}$ be the partial order on heads, $H_i$, of intrinsic sets, $S_i$, in $\mathcal{G}$ such that $H_i \prec_{\mathcal{I}(\mathcal{G})} H_j$ whenever $S_i \subset S_j$.*

**Theorem 14** *The nested ingenuous parameterization of an ADMG $\mathcal{G}$ with nodes $V$ parameterizes precisely those distributions $p(x_V)$ obeying the nested global Markov property with respect to $\mathcal{G}$.*

---
[2]This kernel, viewed causally, is $p(x_2, x_4|do(x_1, x_3))$.

*Proof:* Let $\prec_{\mathcal{I}(\mathcal{G})}$ be the partial ordering on heads given in Definition 13. We proceed by induction on this ordering. For the base case, we know that singleton heads $\{h\}$ with empty tails are parameterized by $\lambda_h^h$. If a singleton head has a non-empty tail, the conclusion follows immediately by Lemma 12.

Now, suppose that we wish to find the kernel with a non-singleton head $H^\dagger$ and a tail $T^\dagger$ corresponding to the intrinsic set $S^\dagger$. Assume, by inductive hypothesis, that we have already obtained the kernels with all heads $H$ such that $H \prec_{\mathcal{I}(\mathcal{G})} H^\dagger$. We claim this is sufficient to give the $(|H^\dagger|-1)$-dimensional marginal kernels $q_{S^\dagger}(x_{H^\dagger \setminus \{v\}}|x_{T^\dagger})$, for all $v \in H^\dagger$.

Fix a particular $v \in H^\dagger$. The set $S^\dagger \setminus \{v\}$ is reachable, since $V \setminus S^\dagger$ is a set with a valid fixing sequence, and any $v \in H^\dagger$ has no children in $S^\dagger$ in $\phi_{V \setminus S^\dagger}(\mathcal{G})$ so is fixable in $\phi_{V \setminus S^\dagger}(\mathcal{G})$. Theorem 7 and Theorem 8 imply that for every reachable set $A$, (2) holds. Hence:

$$q_{S^\dagger}(x_{S^\dagger \setminus \{v\}}|x_{V \setminus (S^\dagger \setminus \{v\})}) = \prod_{\substack{H \in [\![S^\dagger \setminus \{v\}]\!]_{\mathcal{G}} \\ S: \text{rh}_{\mathcal{G}}(S) = H}} q_S(x_H|x_{\text{tail}(H)}). \quad (4)$$

For any $S$ such that $\text{rh}_{\mathcal{G}}(S) = H$, and $H \subseteq S^\dagger \setminus \{v\}$, $H \prec_{\mathcal{I}(\mathcal{G})} H^\dagger$, hence by the induction hypothesis, the kernel $q_S(x_S|x_{pa(S) \setminus S})$ is already obtained, and all kernels which appear in (4) can be derived by conditioning from some such $q_S(x_S|x_{pa(S) \setminus S})$. The desired kernel $q_{S^\dagger}(x_{H^\dagger \setminus \{v\}}|x_{T^\dagger})$ can itself be obtained from $q_{S^\dagger}(x_{S^\dagger \setminus \{v\}}|x_{\text{pa}(S^\dagger) \setminus S^\dagger})$ by conditioning.

We can repeat this argument for any $v \in H^\dagger$. Finally, the nested ingenuous parameterization contains $\lambda_A^{H^\dagger \cup T^\dagger}$ for $H^\dagger \subseteq A \subseteq H^\dagger \cup T^\dagger$. The result then follows by Lemma 12. □

## 7 Simulations

To illustrate the utility of setting higher order parameters to zero ('removing'), we present a simulation study based on the ADMG in Fig. 5 (b). This graph is a special case of two bidirected chains of $k$ vertices each, with a path of directed edges alternating between the chains, for $k = 4$. The number of parameters in the relevant binary nested Markov model grows exponentially with $k$ in graphs of this type.

Consider also the latent variable model defined by replacing each bidirected edge with an independent latent variable shown in Fig. 5 (a), so that $1 \leftrightarrow 3$ becomes $1 \leftarrow 9 \rightarrow 3$. If the state space of each latent variable is the same and fixed, then the number of parameters in this hidden variable DAG model grows only linearly in $k$. This suggests that the nested Markov model may include higher order parameters which are of actual increases in deviance looks much like the relevant $\chi^2$-distribution if we remove interactions of order 6 and higher. The third histogram shows that this starts to break down slightly when 5-way interactions are also zeroed.

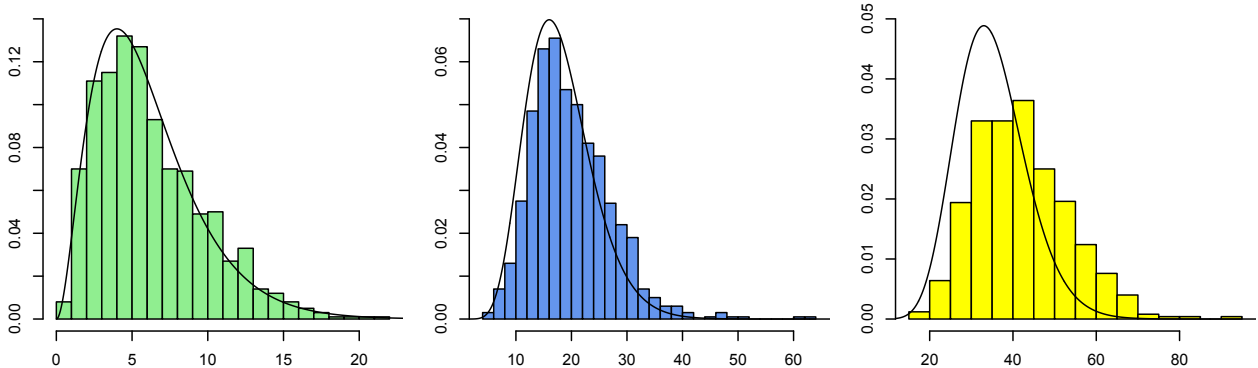

Figure 4: Histograms showing the increase in deviance associated with setting to zero any nested log-linear parameters with effects higher than orders (from left to right) seven, six and five respectively. This corresponds to removing 6, 18 and 35 parameters respectively; the relevant $\chi^2$ density is plotted in each case.

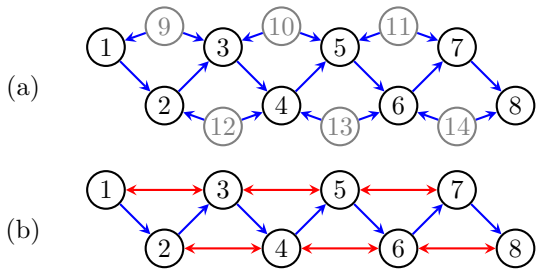

Figure 5: (a) A hidden variable DAG used to generate samples for Section 7. (b) The latent projection of this generating DAG.

These results suggest that higher order parameters are often not useful for explaining finite datasets, and more parsimonious models can be obtained by removing them; a similar simulation was performed for the Markov case in [6].

### 7.1 Distinguishing Graphs

The use of score-based search methods for recovering nested Markov models had been investigated [15]. It was found that relatively large sample sizes were required to reliably recover the correct graph, even in examples with only 4 or 5 binary nodes and after ensuring that the underlying distributions were approximately faithful to the true graph. One phenomenon identified was that incorrect but more parsimonious graphs, especially DAGs, tended to have lower BIC scores than the correct models, which include higher order parameters. Although BIC is guaranteed to be smaller on the correct model asymptotically, in finite samples it applies strong penalties for having additional parameters with little explanatory power.

Here we present a simulation to show how the new parameterization can help to overcome this difficulty. Using the method described in the previous subsection, we generated 1,000 multivariate binary distributions which were nested Markov with respect to the graph in Fig. 1 (b). For each distribution we generated a dataset, and fitted the data to the correct model, which has 16 parameters, as well as the two DAGs given in Fig. 7 (a) and (b), which each have not really necessary in this case (though the higher order parameters may become necessary again if the state space of latent variables grows).

We generated distributions from the latent variable model associated with the DAG in Fig. 5 (a) as follows: each of the six latent variables takes one of three states with equal probability, and each observed variable takes the value 0 with a probability generated as an independent uniform random variable on $(0, 1)$, conditional upon each possible value of its parents.

For each of 1,000 distributions produced independently using this method, we generated a dataset of size 5,000. We then fitted the nested model generated by the graph in Fig. 5 (b) to each dataset by maximum likelihood estimation, using a variation of an algorithm found in [5], and measured the increase in deviance associated with zeroing any nested ingenuous parameters corresponding to effects above a certain order. If these parameters were truly zero, we would expect the increase to follow a $\chi^2$-distribution with an appropriate number of degrees of freedom; the first two histograms in Fig. 4 demonstrate that the distribution

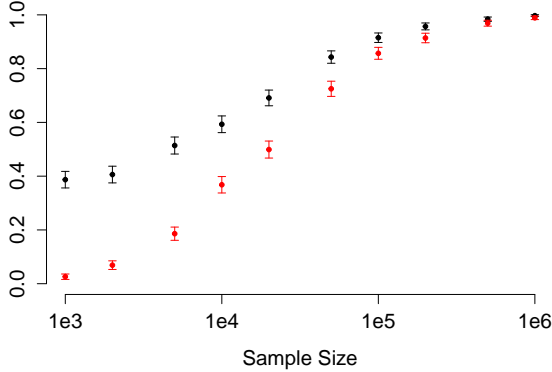

Figure 6: From the experiment in Section 7.1: in red, the proportion of times graph in Fig. 1 (b) had lower BIC than the DAGs in Fig. 7, for varying sample sizes; in black, the proportion of times some restricted version of this model had a lower BIC than any restricted versions of either DAG.

11 parameters. This was repeated at various sample sizes.

The plot in Fig. 6 shows, in red, the proportion of times in which the BIC score for the correct model was lower than that for each of the DAGs, at various sample sizes. The correct graph only has the lowest BIC score of the three graphs on less than 3% of runs at sample size of $n = 1{,}000$, increasing to around 50% for $n = 20{,}000$.

In addition to the full models, we fitted the datasets to versions of the models with higher order parameters removed; the graph in Fig. 1 (b) can be restricted by zeroing the 5-way parameter (leaving 15 free parameters), the 4-way and and above (13 params), or 3-way and above (10 params). Similarly we can restrict the DAGs to have no 3-way effects, giving each model 10 free parameters. Fig. 6 shows, in black, the proportion of times that one of these restricted versions of the true model had a lower BIC than any version of either DAG model. We see that the correct graph has the lowest score in 40% of runs at $n = 1{,}000$, rising to around 70% at $n = 20{,}000$. Note that these results should not be compared directly to those in [15], since the single ground truth law used in that paper was generated so as to ensure faithfulness to the correct graph, whereas we are randomly sampling multiple laws without bothering to ensure any particular properties in these laws other than consistency with the underlying DAG.

These results suggest that these submodels of the nested model may be advantageous in recovering the correct graphical structure using score-based methods.

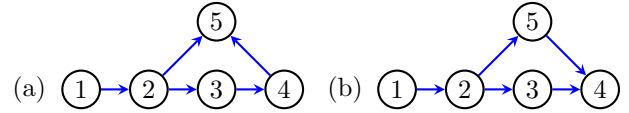

Figure 7: (a) and (b) two DAGs with the same skeleton as the graph in Fig. 1 (b).

Note that determining which higher order parameters should be set to zero for a given data set and sample size remains non-trivial. Automatic selection might be possible with an $L_1$-penalized approach [16, 4].

## 8 Discussion and Conclusions

We have introduced a new log-linear parameterization of nested Markov models over discrete state spaces. The log-linear parameters correspond to 'interactions' in kernels obtained after an iterative application of truncation and marginalization steps (informally 'interactions in interventional densities'). By contrast the Möbius parameters [15] correspond to context specific effects in kernels (informally 'context specific causal effects').

We have shown by means of a simulation study that in cases where data is generated from a marginal of a DAG with 'weak confounders', we can reduce the dimension of the model by ignoring higher order interaction parameters, while retaining the advantages of nested Markov models compared to modeling weak confounding directly in a DAG.

Though there is no efficient, closed form mapping from ingenuous parameters to either Möbius parameters or standard probabilities, this is a smaller disadvantage than it may seem. This is because in cases where the ingenuous parameterization was used to select a particular submodel based on a dataset, we may still reparameterize and use Möbius parameters, or even standard joint probabilities if desired. Moreover, this reparameterization step need only be performed once, compared to multiple calls to a fitting procedure which identified the particular graph corresponding to our submodel in the first place.

The ingenuous and the Möbius parameterizations are thus complementary. The natural application of the ingenuous parameterization is in learning graph structure from data in situations where many samples are not available, but we expect most confounding to be weak. The natural application of the Möbius parameterization is the support of probabilistic and causal inference in a particular graph [14, 15], in cases where an efficient mapping from parameters to joint probabilities is important.